\documentclass[10pt,twocolumn,letterpaper]{article}
\ifdefined\pdfminorversion\pdfminorversion=7\fi
\usepackage{physresponse-layout}
\usepackage{amsmath,amssymb,booktabs,graphicx}
\usepackage[T1]{fontenc}
\definecolor{linkblue}{rgb}{0.21,0.49,0.74}\usepackage[pagebackref,hyperfootnotes=false,breaklinks,colorlinks,allcolors=linkblue]{hyperref}
\hypersetup{hypertexnames=false}
\usepackage{array,tabularx,microtype}
\usepackage{placeins}
\usepackage{pifont}
\newcommand{\cmark}{\ding{51}}\newcommand{\xmark}{\ding{55}}
\newcommand{\tablestyle}{\small\renewcommand{\arraystretch}{1.15}\setlength{\tabcolsep}{4pt}}

\title{PhysResponse: Probing the Physics of\\ Video Generation with Controlled Interventions}
\author{%
Yuhan Li\textsuperscript{1,*}\quad Shuran Zhang\textsuperscript{1,*}\quad Haixin Tian\textsuperscript{1}\quad Baonuo Rao\textsuperscript{1}\quad Weipeng Li\textsuperscript{1}\\
Yujia Hou\textsuperscript{1}\quad Zhenyu Wang\textsuperscript{1}\quad Zijun Wang\textsuperscript{1}\quad Luoyang Lin\textsuperscript{1}\quad Lijing Luo\textsuperscript{1}\\
Jiehui Huang\textsuperscript{2}\quad Kaiyuan Yang\textsuperscript{1}\quad Jingbin Tan\textsuperscript{1}\quad Panwen Hu\textsuperscript{3,\ensuremath{\dagger}}\quad Xiaodan Liang\textsuperscript{1,3,\ensuremath{\dagger}}\\[8pt]
\textsuperscript{1}Sun Yat-sen University\qquad \textsuperscript{2}The Hong Kong University of Science and Technology\\
\textsuperscript{3}Mohamed bin Zayed University of Artificial Intelligence}
\date{}
\renewcommand{\paragraph}[1]{\vspace{.5em}\noindent\textbf{#1}}
\makeatletter
\def\@maketitle{%
  \nointerlineskip
  \vbox to 138.27pt{%
    \centering
    \vskip -12pt
    \hrule height 2.4pt
    \vskip 12pt
    {\Large\bfseries\@title\par}%
    \vskip 10pt
    \hrule height 0.8pt
    \vskip 14.8pt
    {\large\begin{tabular}{@{}c@{}}\@author\end{tabular}\par}%
    \vfil
  }%
}
\makeatother
\begin{document}
\twocolumn[{\csname @maketitle\endcsname
\centering
\includegraphics[width=\linewidth]{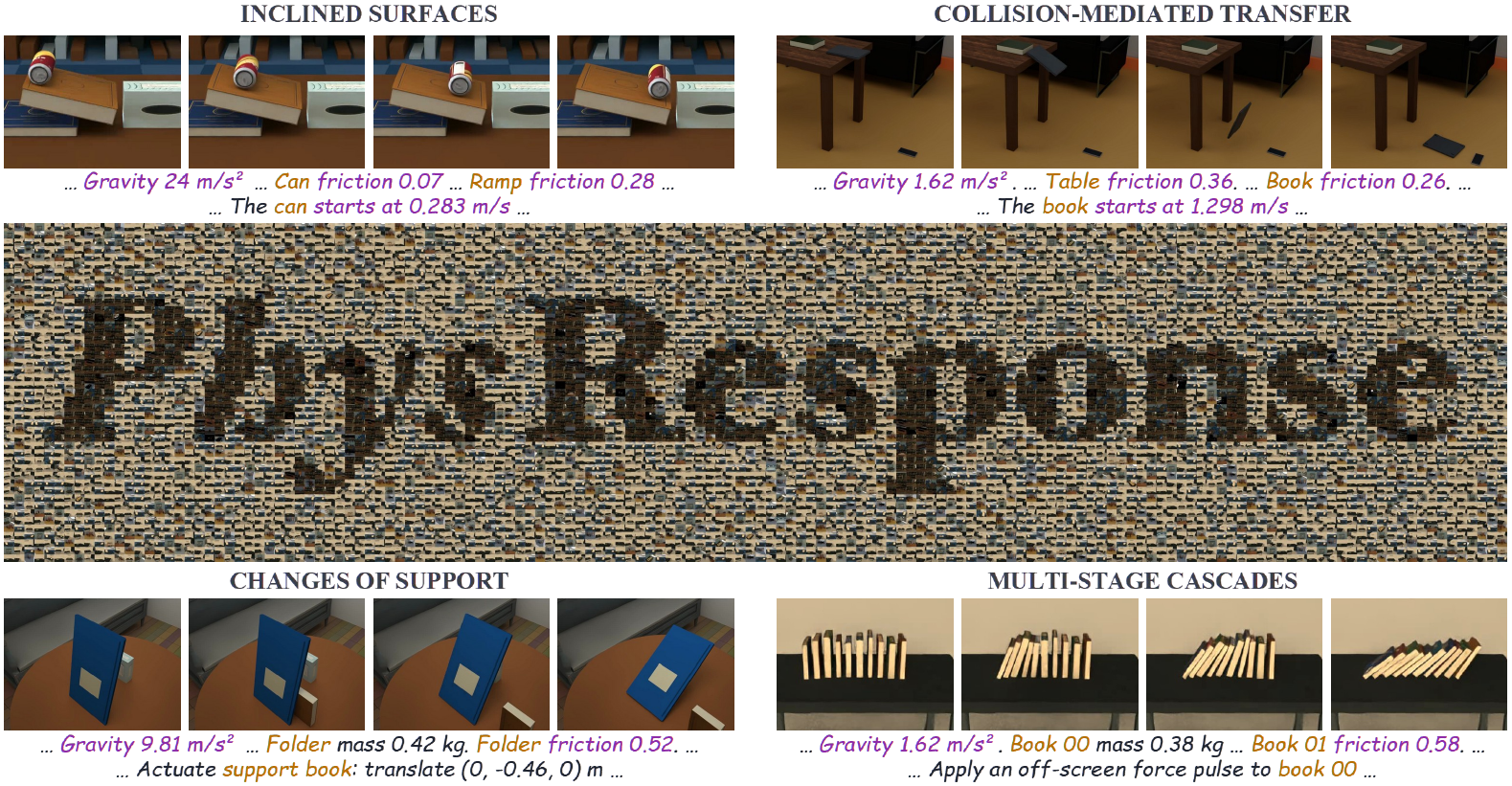}
\captionsetup{hypcap=false}
\captionof{figure}{We present \textbf{PhysResponse}, 72,000 simulated videos of 20 multi-object interactions with set physical conditions, and a benchmark that recovers gravity, friction and speed from generated videos. One video and prompt excerpt per interaction family.}
\label{fig:teaser}\vspace{8.032pt}
\captionsetup{hypcap=true}
}]

\begingroup
\renewcommand{\thefootnote}{}
\footnotetext{\textsuperscript{*}Equal contribution.\quad
\textsuperscript{\ensuremath{\dagger}}Co-corresponding authors.\par
\href{mailto:liyh693@mail2.sysu.edu.cn}{\nolinkurl{liyh693@mail2.sysu.edu.cn}}}
\endgroup
\begin{abstract}
Video generation models are increasingly viewed as world simulators, which requires not only plausible motion but dynamics that respond correctly to specified physical parameters. Existing models, datasets and benchmarks, however, largely target physical plausibility, leaving this physical parameter responsiveness insufficiently explored. We introduce \textbf{PhysResponse}, a benchmark that measures it at two levels: scene-conditioned simulator inversion recovers the effective gravity, friction and initial speed of generated motion, and event evaluation tests outcomes near and across critical transitions. Evaluating ten image-to-video models, we find that even the lowest parameter errors exceed those of physically correct videos by over an order of magnitude, that models rarely reproduce both outcomes of a transition pair, and that the two levels expose complementary failures that visual metrics such as PSNR miss. To supply the missing training signal, we construct 72,000 simulated videos of 20 multi-object interaction types with controlled parameters; fine-tuning on them lowers parameter errors, even on held-out interaction types, and transfers to real videos and an external physics benchmark. Continued training with a numerical encoder further lowers all three errors on held-out types. PhysResponse provides a foundation for world simulators that follow physical conditions, not just appearances.
\par\end{abstract}
\section{Introduction}
\label{sec:motivation_revision}

Rapid advances in video generation~\cite{cogvideox, hunyuanvideo15, wan, minimaxh3} have enabled unprecedented visual fidelity and realism, motivating a growing body of research to develop video generation models into world simulators~\cite{cosmospredict2,cosmos3,physalign,physrvg} capable of understanding and predicting real-world physical dynamics. Alongside these advances, physics-focused datasets~\cite{phyco,phyparam} and benchmarks have emerged to improve and evaluate models\textquoteright{} ability to generate physically plausible motion and interactions. Together, these efforts have established physical plausibility as an important criterion for assessing progress toward world simulation.

However, physical plausibility alone is insufficient for a world simulator. Beyond producing realistic motion, a simulator must capture how dynamics change in response to specified physical parameters. In embodied AI, for example, it should predict how variations in gripper friction affect object manipulation. Such capabilities are essential for simulating interactions across scenarios and physical conditions. Yet existing models~\cite{physalign,videogpa}, datasets~\cite{physinone,phyparam}, and benchmarks~\cite{physicsiq, rigidbench,morpheus} largely focus on a common question: \textit{ Do the generated outcomes appear consistent with physical common sense?} This leaves a more precise question insufficiently explored: \textit{Do the generated dynamics respond correctly to the specified physical parameters?} We refer to this capability as physical parameter responsiveness. Addressing it requires evaluating not only whether a video depicts plausible dynamics, but also whether those dynamics are consistent with the particular parameter values provided.

\begin{figure}[t]
\centering
\includegraphics[width=\linewidth]{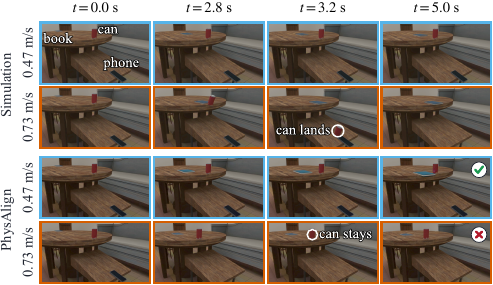}
\caption{\textbf{Speed decides the outcome in simulation but often not in generation.} In simulation only the faster book tips the can onto the shelf, but PhysAlign, even told so, leaves it on the table at both speeds. \textcolor[HTML]{1A9850}{\cmark}/\textcolor[HTML]{B2182B}{\xmark}: match/contradict the simulation.}
\label{fig:response_motivation}
\end{figure}

To address this gap, we first introduce a benchmark and evaluation protocol for measuring physical parameter responsiveness at both the event and parameter levels. Using a physics simulator, we render multi-object interactions across 20 scenarios under varying combinations of gravitational acceleration, friction, and initial speed. Our event-level metrics assess whether generated videos exhibit the expected events under those conditions, while our parameter-level metrics measure how closely physical parameters recovered from the generated dynamics match their specified values. Together, these metrics evaluate both event outcomes and quantitative consistency with the conditioning parameters. Our evaluation shows that current models struggle to respond accurately to specified physical parameters (Fig.~\ref{fig:response_motivation}).

One factor that may contribute to this limitation is the lack of training data that systematically captures how physical parameters shape observable dynamics. We therefore develop a data generation pipeline with control over physical parameters and construct a dataset of 72,000 videos spanning 20 scenarios. The dataset depicts object motion and interactions under diverse parameter combinations, with particular emphasis on capturing variations that produce distinct motion patterns or event outcomes. This diversity provides training examples that connect parameter values to their observable physical effects. Experiments show that fine-tuning on our dataset improves physical parameter responsiveness, supporting the value of targeted data for learning dynamics.

Beyond data, we investigate how model design can further improve physical parameter responsiveness. Specifically, we augment the baseline model with a numerical encoder and fine-tune the resulting model on our dataset. Continued training with the numerical encoder further improves parameter responsiveness, lowering all three errors on held-out interaction types, suggesting that strengthening numerical conditioning is a promising direction for enabling world simulators to accurately reproduce dynamics under specified physical conditions. Taken together, our benchmark, dataset, and model exploration provide a foundation for studying world simulation beyond general physical plausibility. 
\par\noindent Our contributions are threefold:
\begin{itemize}
    \item  A \textbf{benchmark} for physical parameter responsiveness. We introduce a benchmark spanning 20 scenarios, together with event-level and parameter-level metrics that assess whether generated dynamics correctly reflect specified physical parameters. Our evaluation identifies substantial limitations in current state-of-the-art models.
    \item  A \textbf{dataset} with controlled physical parameters. We develop a configurable simulation pipeline and construct a dataset of 72,000 videos covering diverse parameter combinations and their resulting dynamics. We show that fine-tuning on this dataset improves physical parameter responsiveness.
    \item An \textbf{investigation} of numerical conditioning. We augment a baseline model with a numerical encoder and show that continued training with it further improves parameter responsiveness on held-out interaction types, highlighting numerical conditioning as a promising direction for parameter-controlled world simulation.
\end{itemize}

\section{Related Work}
\label{sec:related}

\begin{figure*}[t]
\centering
\scalebox{1}[1.10]{\includegraphics[width=\textwidth]{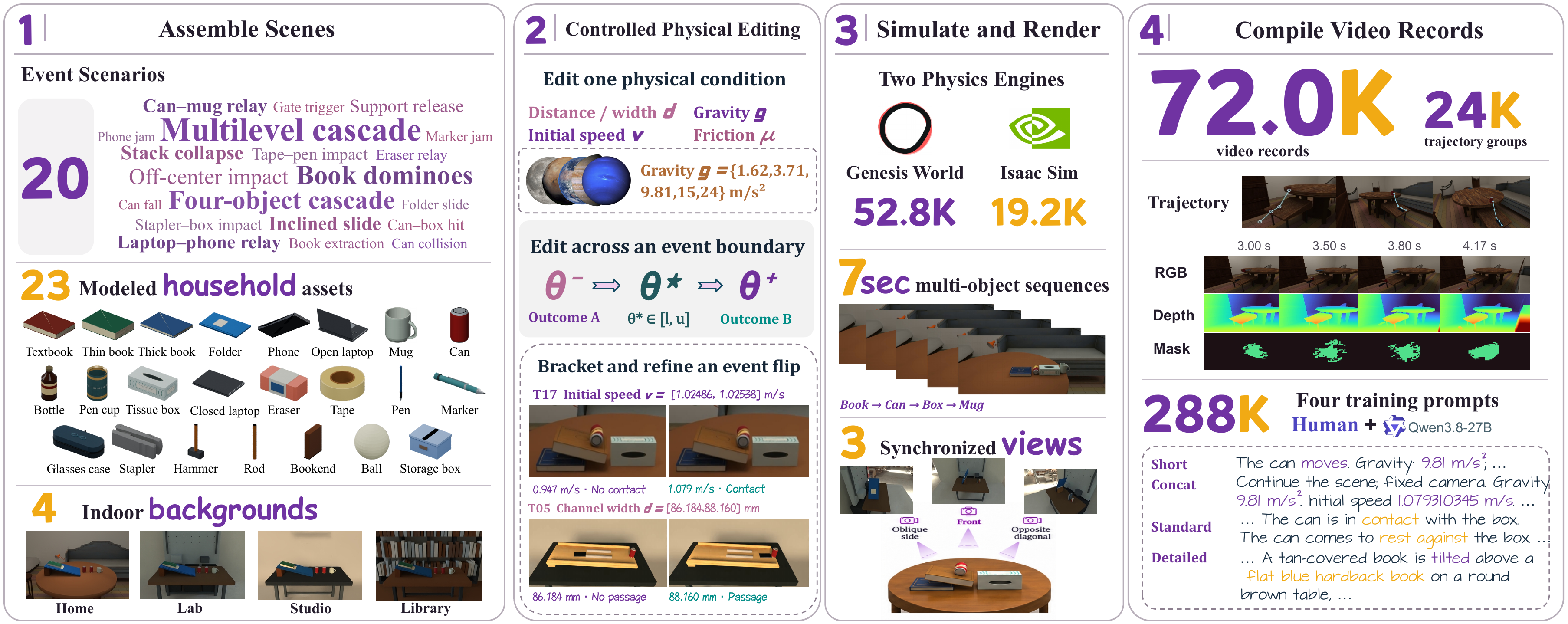}}
\caption{\textbf{Constructing PhysResponse.} Scenes are assembled from household assets and four backgrounds, edited one physical condition at a time, simulated in Genesis and Isaac and rendered from three views; panel 2 brackets two reference transitions, in the initial speed of T17 and the channel width of T05. Depth and masks are derived from RGB.}
\label{fig:construction_pipeline}
\end{figure*}
\paragraph{Physical conditioning for video generation.}
Video generators use simulated dynamics and explicit controls to specify physical behavior~\cite{physgen,physctrl,forceprompting,newtongen,psivg}, while physics-oriented post-training improves motion realism~\cite{pisa,newtonrewards,physrvg}. PhyCo~\cite{phyco} introduces controllable physical priors; PhyParam~\cite{phyparam} pairs simulated rigid-body videos with object-level parameters and evaluates physical-law consistency. PhysResponse focuses on multi-object interactions, linking controlled physical inputs to both the motion they produce and the events they trigger.

\paragraph{Physical evaluation of video models.}
Benchmarks assess physical plausibility through human or vision-language judgments~\cite{videophy,videophy2,phygenbench,worldmodelbench,vbench2,phyworldbench,t2vphysbench,applepi}, or measure motion against real and simulated references~\cite{physicsiq,rigidbench,gauge,morpheus}. PhysInOne~\cite{physinone} measures spectral motion agreement, WorldBench~\cite{worldbench} examines numerical physical properties, and Principia~\cite{principia} tests inter-object relations. Fall-time analyses~\cite{slower} and PhysWeep~\cite{physweep} recover physical quantities from generated motion, with PhysWeep studying sweeps across five single-object systems. Counterfactual benchmarks test responses to changed conditions in reasoning~\cite{cophy,clevrer,comphy,physion,physionpp} and generation~\cite{whatifworld,cronos,causalcascade}: PhysicsLENS~\cite{physicslens} varies stated physical properties, and GeoFlip~\cite{geoflip} tests outcome flips under geometric interventions. PhysResponse brings parameter recovery and transition-based event evaluation together in multi-object scenes, testing whether generated motion matches the specified conditions and produces the corresponding interactions.

\paragraph{Analysis by synthesis.}
Physical inference matches simulations to observed motion~\cite{galileo,videophysedit}, learns representations from video~\cite{physics101}, or uses differentiable simulation~\cite{pacnerf}; related work probes dynamic properties in video foundation models~\cite{idpp}. We use simulation matching to recover effective gravity, friction and initial speed, conditioning on known geometry, camera, initial state and non-target settings. This provides a quantitative account of the dynamics behind generated interactions.

\begin{figure}[!t]
\centering
\includegraphics[width=\linewidth]{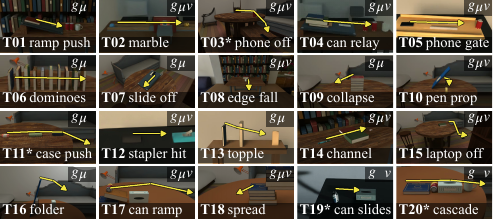}
\caption{\textbf{Twenty everyday interaction types.} Arrows: motion; letters: parameters evaluated; $^*$held out of training.}
\label{fig:dataset_overview}
\end{figure}

\section{The PhysResponse Dataset}
\label{sec:dataset}
\subsection{Interaction Coverage and Physical Conditions}

PhysResponse links editable physical conditions to multi-object motion and interaction outcomes (Fig.~\ref{fig:dataset_overview}; Table~\ref{tab:dataset_scale}). Its 20 everyday interaction types (scenarios T01--T20) span four mechanisms: inclined-surface motion, collision-mediated transfer, changes of support, and multistage cascades. Gravity affects acceleration and fall timing; friction affects sliding and motion transfer through contact; initial speed determines whether and when objects reach one another.

The dataset contains 72,000 videos from 24,000 simulation runs, each rendered from three synchronized viewpoints. Each type contributes 3,600 videos: 2,640 from Genesis~\cite{genesis} and 960 from Isaac Sim~\cite{isaacsim}. All views of a run share the same data split.

\begin{table}[t]
\caption{\textbf{Datasets and evaluation capabilities.} Recovery: parameters recovered from generated videos; Ctrl.: one parameter varied; flips: opposite reference outcomes. R/S: real/simulated; P: prompts; proc.: on demand; (\cmark): partial; $^\dagger$by material; $^\S$geometry only. $e,\rho,\eta,k,m,F$: restitution, density, viscosity, stiffness, mass, force.}
\label{tab:dataset_scale}
\centering\tablestyle\footnotesize\setlength{\tabcolsep}{0.7pt}
\newcommand{\hd}[1]{{\scriptsize #1}}
\begin{tabular}{@{}lr@{\,}lccccccc@{}}
\toprule
& & & & \hd{Scene} & \multicolumn{2}{c}{\hd{Recovery}} & & & \\
\cmidrule(lr){5-5}\cmidrule(lr){6-7}
Work & \multicolumn{2}{c}{Videos} & Set param. & \hd{Multi} & \hd{Any} & \hd{Multi} & \hd{Ctrl.} & \hd{Flips} & \hd{Train}\\
\midrule
Physics-IQ~\cite{physicsiq}      & 396   & R     & \xmark              & \cmark   & \xmark & \xmark & \xmark   & \xmark        & \xmark\\
PhyGenBench~\cite{phygenbench}   & 160   & P     & \xmark              & (\cmark) & \xmark & \xmark & \xmark   & \xmark        & \xmark\\
VideoPhy-2~\cite{videophy2}      & 3.9K  & P     & \xmark              & \cmark   & \xmark & \xmark & \xmark   & \xmark        & \xmark\\
WorldBench~\cite{worldbench}     & 748   & R{+}S & $\mu,\eta^\dagger$  & \cmark   & \cmark & \xmark & \xmark   & \xmark        & \xmark\\
PhysInOne~\cite{physinone}       & 2M    & S     & $\mu,e,\rho,\ldots$ & \cmark   & \xmark & \xmark & \xmark   & \xmark        & \cmark\\
PhyParam~\cite{phyparam}         & 130K  & S     & $g,\mu,m,F,e$       & \cmark   & \xmark & \xmark & (\cmark) & \xmark        & \cmark\\
What-If World~\cite{whatifworld} & 638   & P     & \xmark              & \cmark   & \xmark & \xmark & \cmark   & \xmark        & \xmark\\
RigidBench~\cite{rigidbench}     & 5.1K  & S     & \xmark              & \cmark   & \xmark & \xmark & \xmark   & \xmark        & \cmark\\
PhysWeep~\cite{physweep}         & proc. & S     & $g,\mu,e,k,\ldots$  & \xmark   & \cmark & \xmark & \cmark   & \xmark        & \xmark\\
GeoFlip~\cite{geoflip}           & 964   & S     & \xmark              & \cmark   & \xmark & \xmark & \cmark   & (\cmark)$^\S$ & \xmark\\
\midrule
\textbf{PhysResponse}            & 72K   & S     & $g,\mu,v$           & \cmark   & \cmark & \cmark & \cmark   & \cmark        & \cmark\\
\bottomrule
\end{tabular}
\end{table}

\subsection{Construction}

Scenes specify object configurations, initial states, contact settings, and prescribed controls (Fig.~\ref{fig:construction_pipeline}). We edit gravity across five levels from 1.62 to 24\,m/s$^2$, with scene-specific friction and initial-speed settings. Four backgrounds and three viewpoints provide appearance variation. Available simulation states and contact records support parameter measurements and reference event outcomes.

Controlled interventions vary one target condition while fixing non-target conditions. Dependent quantities change accordingly, such as angular speed when editing a rolling object's linear speed. Reference simulations locate and refine parameter intervals across which event outcomes flip. Configurations resampled across multiple inputs populate the training pool, providing examples of how physical conditions shape motion and interactions.

Core100 selects five cases per interaction type for evaluation. Figure~\ref{fig:construction_pipeline} illustrates training prompts; evaluation uses four English formats: \emph{Short} states the interaction and physical values; \emph{Standard} embeds these values in a natural description; \emph{Detailed} adds the initial spatial arrangement; and \emph{Concat} appends values to a fixed template. Some prompts also state expected event outcomes. These resources support both learning physical responses and evaluating their motion-level parameters and event-level consequences.

\section{Parameter-Level Evaluation}
\label{sec:conditional_reading}

We recover effective parameters by replaying the source scene, varying $\theta\in\{g,\mu,v\}$ while fixing context $c$: geometry, camera, non-target initial and physical conditions, and prescribed controls. Matching replay $\mathcal S(c,\theta)$ to generated motion gives
\begin{equation}
\widehat\theta(c)=\arg\min_{\theta\in\Theta_c}\min_{\Delta\in\mathcal D} L(\theta,\Delta).
\label{eq:conditional_inversion}
\end{equation}
Here $\Delta$ is a bounded onset offset; $\widehat\theta(c)$ explains the motion under the fixed context (Fig.~\ref{fig:identifiability}).

\paragraph{Conditional identifiability.}
Conditioning is essential because different parameters can explain the same motion. In horizontal sliding,
\begin{equation}
a=-\mu g,
\label{eq:sliding_ambiguity}
\end{equation}
so deceleration identifies $\mu g$, not its factors. Fixing friction permits a conditional gravity estimate; a sharp minimum alone cannot establish joint identifiability. Gravity requires gravity-driven motion, friction visible sliding contact, and speed an observable onset with known direction. The loss curve distinguishes localized estimates, one-sided bounds and unidentifiable cases.

\subsection{Observation and simulation matching}
SAM~3.1~\cite{sam3} tracks condition-frame objects. We align backgrounds to compensate camera motion, map frames to physical time, and register projected replay boxes after removing condition-frame offsets. Candidate grids exclude the requested value.

For an observed box $\widehat b_o$ and a projected box $b_o$, the four-edge matching loss is
\begin{equation}
\ell_o(t;\theta,\Delta)=
\frac{\sum_{j=1}^{4}\rho\!\left(\frac{\widehat b_{o,j}(t)-b_{o,j}(t+\Delta;c,\theta)}{d_o}\right)}
{2(\sigma/d_o)^2}.
\label{eq:box_matching}
\end{equation}
Here $d_o$ is the initial box diagonal, $\rho$ the Huber loss and $\sigma$ the tracking-noise scale. $L$ sums visibility-penalized temporal averages over objects with a shared physical explanation, within windows supported by replay repeatability and reference agreement. A common onset offset aligns all objects.

A quadratic fit in log-parameter refines the grid minimum. Reference-calibrated tolerance $\delta$ defines the indistinguishable range, yielding a point estimate, bound or unidentifiable result according to the loss basin.

\begin{figure*}[t]
\centering
\includegraphics[width=\textwidth]{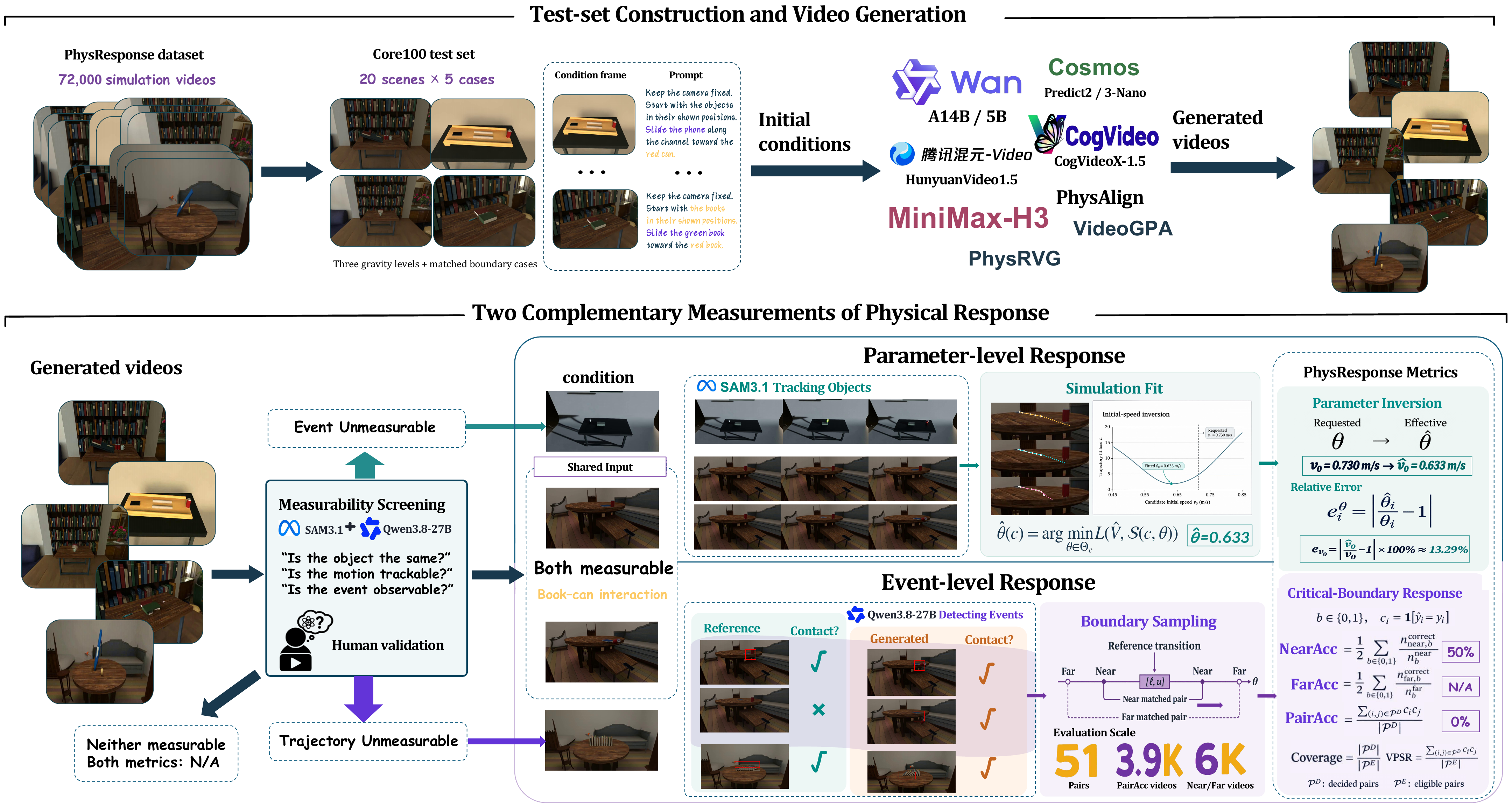}
\caption{\textbf{PhysResponse evaluation overview.} Core100 conditions and prompts produce videos for two complementary measurements. Object tracks support simulator-based parameter recovery; reference transitions and visual event observations support NearAcc, FarAcc and PairAcc. Measurability screening identifies the evidence available to each branch, with human validation of the observations.}
\label{fig:benchmark_pipeline}
\end{figure*}

\subsection{Aggregation and kinematics}
\label{sec:aggregation}
\paragraph{Aggregation.}
For each model and parameter, we rank results within scenes containing at least three results. Stronger parameter constraints rank first; fit residual breaks ties. Parameter error does not affect ranking. We retain the top 10\% (at least one result), pool them with equal total weight per scene, and report the weighted median of $e_i^\theta=|\widehat\theta_i/\theta_i-1|$ for $\theta_i\neq0$. One-sided bounds enter at their grid edge, so their errors are lower bounds. Unmeasurable retained slots count as 100\% error, preventing score gains from excluding difficult samples.

\paragraph{Kinematics.}
Displacement is the generated displacement projected onto the reference direction, divided by reference length: 1 indicates correct motion and 0 no motion. We average over eligible objects. Causal order is the fraction of reference object pairs whose onset order is preserved; nonmoving pairs fail this test.

\begin{figure}[t]
\centering
\includegraphics[width=\linewidth]{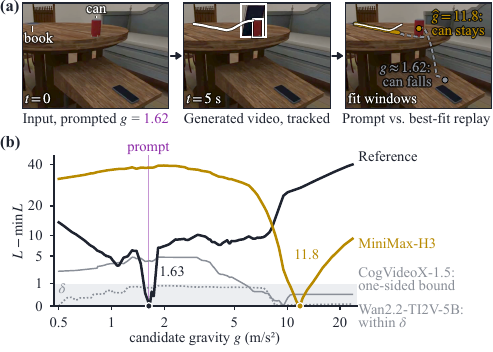}
\caption{\textbf{Recovering gravity from a generated video.} (a)~MiniMax-H3, Moon gravity: tracks (white) match the replay at $\widehat g=11.8$\,m/s$^2$ (gold), not at the prompted $g$ (grey). (b)~Excess fit loss, square-root axis; shaded: tolerance $\delta$; thin grey: a lower bound and an unidentifiable estimate.}
\label{fig:identifiability}
\end{figure}

\section{Event-Level Evaluation}
\label{sec:intervention_events}
Parameter estimates describe generated dynamics; event evaluation tests their interaction consequences. Reference simulations define outcome transitions against which we evaluate individual conditions and paired interventions (Fig.~\ref{fig:benchmark_pipeline}).

\subsection{Object-Centric Event Observation}
\label{sec:object_event}
An event $e$ comprises participating objects, an observable binary predicate $f_e$ and a window $W$, with reference outcome $y_e(c,\theta;W)\in\{0,1\}$. Generated outcomes are evaluated against this reference. SAM~3.1~\cite{sam3} supplies object tracks and local crops; Qwen3.8-27B~\cite{qwen38} examines these alongside full-scene frames.

\subsection{Reference-Defined Critical Transitions}
\label{sec:boundary_construction}
Small changes in physical conditions can initiate contact or redirect a cascade: only the faster book in Fig.~\ref{fig:response_motivation} tips the can onto the shelf. Testing both conditions probes this response.

We fix $c$, $e$ and $W$ and vary $\theta$ along a predefined intervention axis in reference simulation. Opposite reference outcomes at two verified settings define a local transition interval:
\begin{equation}
[l,u]:\quad y_e(c,l;W)\ne y_e(c,u;W),\qquad l<u.
\label{eq:reference_bracket}
\end{equation}
The width $u-l$ gives localization resolution. Intervention axes include physical parameters, geometry and control conditions (Supp.~J).

\subsection{Quantifying Event Responsiveness}
\label{sec:event_metrics}
\paragraph{Single-condition accuracy.}
For setting $\theta$ and the intervention's normalization span $S$, distance to the reference interval is
\begin{equation}
d_{\min}(\theta)=\frac{\max(l-\theta,\theta-u,0)}{S}.
\label{eq:event_dmin}
\end{equation}
Near uses $d_{\min}\leq0.20$, including ties and distance intervals crossing 20\%; Far uses $d_{\min}>0.20$. For each model and prompt, let $a_i$ average decidable correctness at reference event--window point $i$, $N_{G,y}$ contain such points with outcome $y$ in group $G$, and $Y_G$ contain their represented outcomes. For $G\in\{\mathrm{Near},\mathrm{Far}\}$,
\begin{equation}
\mathrm{Acc}_G=\frac{1}{|Y_G|}\sum_{y\in Y_G}
\frac{1}{|N_{G,y}|}\sum_{i\in N_{G,y}}a_i.
\label{eq:event_near_far}
\end{equation}
The tables report this outcome-balanced score when both reference outcomes have decidable observations.

\paragraph{Accuracy on outcome flips.}
PairAcc requires both generated endpoints to match their references. We use 51 video pairs across 18 scenes and 245 event--window registrations, matching scene, event and window; generated endpoints share model, prompt and seed. Let $d_p$ and $c_p$ be the fractions of pair $p$'s targets jointly decidable and jointly correct. Each video pair has unit weight.
\begin{equation}
\begin{gathered}
D_w=\sum_p d_p,\qquad C_w=\sum_p c_p,\\[3pt]
\mathrm{PairAcc}=\frac{C_w}{D_w},\;
\mathrm{Coverage}=\frac{D_w}{51},\;
\mathrm{VPSR}=\frac{C_w}{51}.
\end{gathered}
\label{eq:event_pair_metrics}
\end{equation}
PairAcc measures accuracy among decidable targets; Coverage and VPSR measure decidability and verified success over all 51 pairs.

\begin{table*}[!tp]
\centering
{
\captionsetup{type=table,font=small,skip=4pt}
\caption{\textbf{No model follows the prompted physics.} All four prompt formats pooled (per format: supplement); errors aggregated as in Sec.~\ref{sec:aggregation}. Condition frame: first frame repeated, not ranked. Bold/underlined: best/second.}
\label{tab:main_results}

\centering
\small
\renewcommand{\arraystretch}{1.05}
\setlength{\tabcolsep}{3pt}
\begin{tabular*}{\textwidth}{@{\extracolsep{\fill}}lcccccccccc@{}}
\toprule
& \multicolumn{4}{c}{Visual and motion consistency} & \multicolumn{2}{c}{Kinematics} & \multicolumn{3}{c}{Effective parameter error} & Human\\
\cmidrule(lr){2-5}\cmidrule(lr){6-7}\cmidrule(lr){8-10}\cmidrule(l){11-11}
Model & L2 err.\,$\downarrow$ & S-IoU\,$\uparrow$ & PSNR\,$\uparrow$ & SSIM\,$\uparrow$ & Displ.\,$\to$1 & Order\,$\uparrow$ & $g\downarrow$ & $\mu\downarrow$ & $v\downarrow$ & Meas.\,$\uparrow$\\
\midrule
\textit{Condition frame repeated} & 12.59\% & 0.000 & 37.8 & 0.977 & 0.00 & 0\% & -- & -- & -- & -- \\
\midrule
CogVideoX-1.5 & \underline{13.59\%} & 0.057 & 22.9 & 0.844 & 0.03 & 8\% & 114.5\% & 166.0\% & 93.0\% & 23\% \\
Cosmos-Predict2 & 13.60\% & 0.071 & 20.8 & 0.800 & 0.06 & 8\% & 89.2\% & 76.8\% & 81.9\% & 40\% \\
Cosmos3-Nano & 17.40\% & 0.153 & 29.7 & \underline{0.935} & \underline{0.66} & 23\% & \underline{81.8\%} & 77.0\% & \textbf{53.5\%} & 45\% \\
HunyuanVideo-1.5 & 14.90\% & 0.200 & 27.4 & 0.927 & 0.10 & 26\% & 86.5\% & 77.0\% & 63.9\% & 58\% \\
MiniMax-H3 & 13.93\% & \textbf{0.290} & 27.7 & 0.853 & \textbf{0.80} & \textbf{54\%} & \textbf{73.5\%} & \textbf{68.7\%} & 76.8\% & \textbf{96\%} \\
PhysAlign & 19.29\% & \underline{0.217} & \underline{30.1} & \textbf{0.940} & 0.29 & 27\% & 93.5\% & \underline{74.4\%} & \underline{63.1\%} & \underline{87\%} \\
PhysRVG & \textbf{12.94\%} & 0.049 & \textbf{30.3} & 0.926 & 0.01 & 1\% & 140.3\% & 77.7\% & 93.4\% & 83\% \\
VideoGPA & 16.10\% & 0.064 & 22.0 & 0.806 & 0.11 & 9\% & 86.5\% & 101.1\% & 91.1\% & 29\% \\
Wan2.2-I2V-A14B & 17.16\% & 0.099 & 21.0 & 0.702 & 0.63 & \underline{28\%} & 88.9\% & 74.9\% & 78.4\% & 72\% \\
Wan2.2-TI2V-5B & 16.10\% & 0.051 & 23.2 & 0.819 & 0.06 & 13\% & 94.9\% & 130.3\% & 85.8\% & 35\% \\
\bottomrule
\end{tabular*}
\par}

\vspace{8pt}
{
\captionsetup{type=figure,font=small,skip=4pt}
\centering
\scalebox{1}[1.13]{\includegraphics[width=\textwidth]{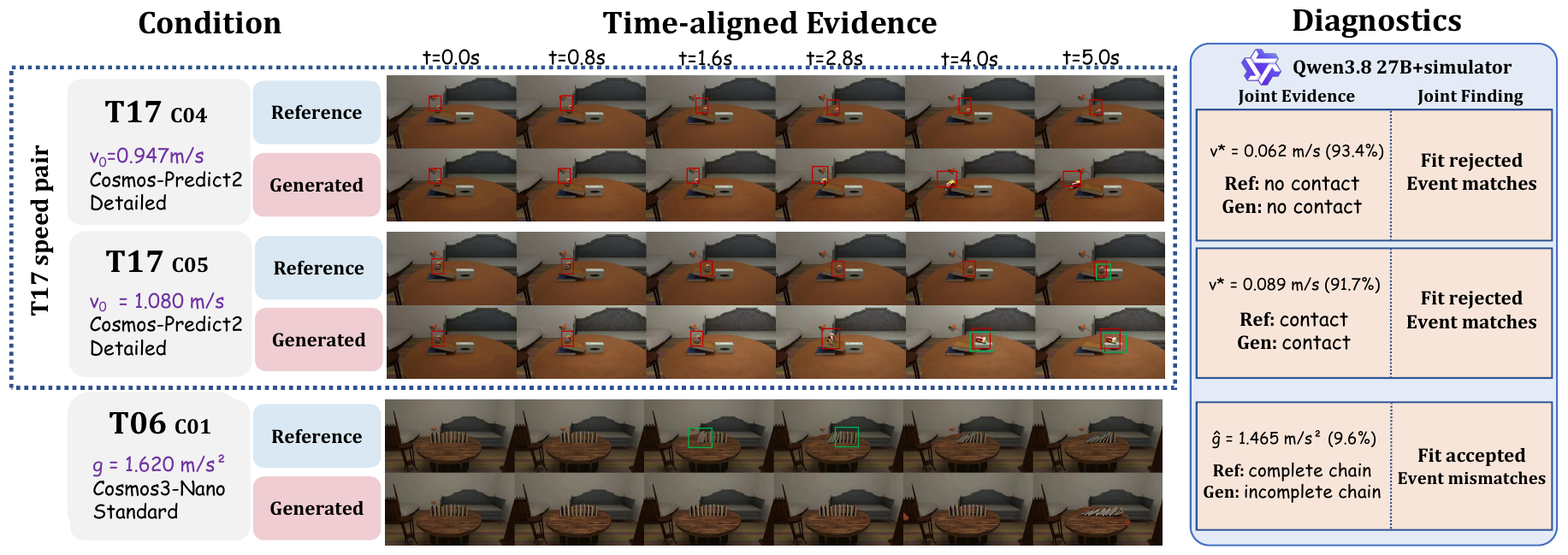}}
\caption{\textbf{Complementarity of parameter- and event-level diagnostics.}
Top: Cosmos-Predict2 matches both contact outcomes, yet both speed fits are rejected; $v^*$ is the lowest-loss candidate, and its percentage a deviation, not an accepted error. Bottom: Cosmos3-Nano recovers gravity within 9.6\% but misses the complete domino chain.}
\label{fig:bidirectional_errors}
\par}

\vspace{8pt}
{
\captionsetup{type=table,font=small,skip=4pt}
\caption{\textbf{Critical-boundary evaluation (\%).}
Standard and Detailed prompts;
metrics: Sec.~\ref{sec:event_metrics};
counts and sensitivity: Supp.~A.}
\label{tab:rq3_strict_pairs}
\centering\small
\setlength{\tabcolsep}{3pt}
\renewcommand{\arraystretch}{1.05}
\begin{tabular*}{\textwidth}{@{\extracolsep{\fill}}lrrrrrrrrrr@{}}
\toprule
 & \multicolumn{5}{c}{Standard (S)} & \multicolumn{5}{c}{Detailed (D)} \\
\cmidrule(lr){2-6}\cmidrule(l){7-11}
Model & NearAcc$\uparrow$ & FarAcc$\uparrow$ & PairAcc$\uparrow$ & Coverage$\uparrow$ & VPSR$\uparrow$ & NearAcc$\uparrow$ & FarAcc$\uparrow$ & PairAcc$\uparrow$ & Coverage$\uparrow$ & VPSR$\uparrow$ \\
\midrule
CogVideoX-1.5 & 45.7 & 50.0 & 0.0 & 67.6 & 0.0 & 43.5 & 54.0 & 4.8 & 82.4 & 3.9 \\
Cosmos-Predict2 & 51.2 & 37.2 & 7.7 & 76.5 & 5.9 & 60.0 & 46.1 & \underline{20.9} & \underline{84.3} & \underline{17.6} \\
Cosmos3-Nano & 58.3 & 50.6 & 12.3 & 63.7 & 7.8 & \underline{62.2} & 45.4 & 14.8 & 79.4 & 11.8 \\
HunyuanVideo-1.5 & 47.7 & 49.4 & 13.7 & 57.2 & 7.8 & 60.8 & \textbf{60.6} & 12.5 & \textbf{86.3} & 10.8 \\
MiniMax-H3 & \textbf{67.0} & 52.9 & \underline{23.1} & \textbf{83.3} & \underline{19.3} & \textbf{69.6} & \underline{60.1} & 18.7 & 82.4 & 15.4 \\
PhysAlign & \underline{60.0} & \underline{60.7} & 20.2 & \underline{82.4} & 16.7 & 45.1 & 45.7 & 11.9 & 82.4 & 9.8 \\
PhysRVG & 50.0 & 52.3 & 0.0 & 64.1 & 0.0 & 50.0 & 50.0 & 0.0 & 66.0 & 0.0 \\
VideoGPA & 54.2 & 52.9 & 7.2 & 54.2 & 3.9 & 53.2 & 56.8 & 15.5 & 63.1 & 9.8 \\
Wan2.2-I2V-A14B & 48.3 & \textbf{62.8} & \textbf{30.7} & 74.5 & \textbf{22.9} & 59.3 & 55.4 & \textbf{42.1} & 82.4 & \textbf{34.6} \\
Wan2.2-TI2V-5B & 35.4 & 25.0 & 0.0 & 36.3 & 0.0 & 40.9 & 55.6 & 8.3 & 70.6 & 5.9 \\
\bottomrule
\end{tabular*}
\par}

\end{table*}

\begin{table*}[!t]
\caption{\textbf{Fine-tuning on PhysResponse lowers all three parameter errors.} Wan2.2-TI2V-5B on the 100 cases; bold: best, underlined: second. Text-only versions: Supp.\ Table~34; event results: Table~\ref{tab:ft51_comparison}.}
\label{tab:training_utility}

\centering
\small
\renewcommand{\arraystretch}{1.05}
\setlength{\tabcolsep}{3pt}
\begin{tabular*}{\textwidth}{@{\extracolsep{\fill}}lcccccccccc@{}}
\toprule
& \multicolumn{4}{c}{Visual and motion consistency} & \multicolumn{2}{c}{Kinematics} & \multicolumn{3}{c}{Effective parameter error} & Human\\
\cmidrule(lr){2-5}\cmidrule(lr){6-7}\cmidrule(lr){8-10}\cmidrule(l){11-11}
Model & L2 err.\,$\downarrow$ & S-IoU\,$\uparrow$ & PSNR\,$\uparrow$ & SSIM\,$\uparrow$ & Displ.\,$\to$1 & Order\,$\uparrow$ & $g\downarrow$ & $\mu\downarrow$ & $v\downarrow$ & Meas.\,$\uparrow$\\
\midrule
Wan2.2-TI2V-5B & 16.10\% & 0.051 & 23.2 & 0.819 & 0.06 & 13\% & 94.9\% & 130.3\% & 85.8\% & 35\% \\
\quad PR-S & \textbf{13.84\%} & \textbf{0.305} & \textbf{32.1} & \textbf{0.949} & \underline{0.59} & \underline{20\%} & \textbf{69.1\%} & \textbf{70.6\%} & \textbf{45.6\%} & \textbf{80\%} \\
\quad PR-D & \underline{14.69\%} & \underline{0.292} & \underline{32.0} & \underline{0.948} & \textbf{0.72} & \textbf{25\%} & \textbf{69.1\%} & \underline{87.6\%} & \underline{67.3\%} & \underline{78\%} \\
\bottomrule
\end{tabular*}
\end{table*}

\section{Experiments}

\subsection{Setup and metric validation}\label{sec:parameter_validation}
We evaluate ten models~\cite{cogvideox,cosmospredict2,cosmos3,hunyuanvideo15,minimaxh3,physalign,physrvg,videogpa,wan} on Core100 using four prompt formats and seed 0: 4,000 generated videos. Videos retain native timing. Parameter and motion scores pool formats; event tables report Standard and Detailed.

\paragraph{Measures.} Alongside parameter and event metrics (Secs.~\ref{sec:aggregation}--\ref{sec:event_metrics}), we report video-averaged PSNR, SSIM~\cite{ssim}, displacement L2 error relative to image height, and moving-region S-IoU~\cite{physicsiq}. Meas. is the proportion of videos judged measurable by blind human annotators.

\paragraph{Metric validation.} Processing GT videos with the same pipeline yields gravity, friction and speed errors of 3.6\%, 1.4\% and 1.1\%; displacement is 0.91 and causal-order accuracy 99\%. These establish a reference for interpreting generated-video errors. Fitting baselines and perturbation controls appear in Supp.~Tables~31--33.

Against blind human annotations of all 4,000 videos, automatic measurability agreement is 74\%. Of 6,070 binary human-labeled events in human-measurable videos, 68\% receive automatic binary judgments, with 79\% agreement (Supp.~Table~26).

\label{tab:rq3_near_far}

\begin{table*}[!t]
\caption{\textbf{Event response after fine-tuning (\%).} Standard and Detailed prompts; baseline and metrics as in Table~\ref{tab:rq3_strict_pairs}. Text-only results: Supp.~Table~17. Bold/underlined: best/second.}
\label{tab:ft51_comparison}
\centering\small
\setlength{\tabcolsep}{3pt}
\renewcommand{\arraystretch}{1.05}
\begin{tabular*}{\textwidth}{@{\extracolsep{\fill}}lrrrrrrrrrr@{}}
\toprule
 & \multicolumn{5}{c}{Standard (S)} & \multicolumn{5}{c}{Detailed (D)} \\
\cmidrule(lr){2-6}\cmidrule(l){7-11}
Model & NearAcc$\uparrow$ & FarAcc$\uparrow$ & PairAcc$\uparrow$ & Coverage$\uparrow$ & VPSR$\uparrow$ & NearAcc$\uparrow$ & FarAcc$\uparrow$ & PairAcc$\uparrow$ & Coverage$\uparrow$ & VPSR$\uparrow$ \\
\midrule
Wan2.2-TI2V-5B & \underline{35.4} & 25.0 & 0.0 & 36.3 & 0.0 & 40.9 & 55.6 & \underline{8.3} & 70.6 & \underline{5.9} \\
\quad PR-S & \textbf{42.8} & \textbf{61.0} & \textbf{15.0} & \textbf{96.1} & \textbf{14.4} & \underline{50.4} & \textbf{60.4} & \textbf{10.2} & \textbf{96.1} & \textbf{9.8} \\
\quad PR-D & 33.0 & \underline{58.0} & \underline{14.3} & \textbf{96.1} & \underline{13.7} & \textbf{54.7} & \underline{60.2} & 4.8 & \textbf{96.1} & 4.6 \\
\bottomrule
\end{tabular*}
\end{table*}

\begin{table}[!t]
\caption{\textbf{Fine-tuning on PhysResponse transfers to PhysInOne.} 100 training-split scenes, seed 0; $^\dagger$fine-tuned by the PhysInOne authors on this split. Bold/underlined: best/second.}
\label{tab:physinone_main}

\centering\small\renewcommand{\arraystretch}{1.05}
\setlength{\tabcolsep}{2pt}
\begin{tabular*}{\linewidth}{@{\extracolsep{\fill}}lrrrrr@{}}
\toprule
Model & PMF\,$\uparrow$ & PSNR\,$\uparrow$ & SSIM\,$\uparrow$ & LPIPS\,$\downarrow$ & FVD\,$\downarrow$\\
\midrule
Wan2.2-TI2V-5B & 2.058 & 12.5 & 0.404 & 0.503 & 1334.3\\
\quad PhysInOne LoRA$^\dagger$ & 2.835 & 15.8 & 0.622 & 0.337 & 893.4\\
\quad PhysInOne SFT$^\dagger$ & 2.880 & 15.7 & 0.685 & 0.305 & \textbf{772.3}\\
\midrule
\quad PR-S & \underline{3.144} & \textbf{17.8} & \underline{0.738} & \underline{0.239} & 794.7\\
\quad PR-D & \textbf{3.148} & \textbf{17.8} & \textbf{0.740} & \textbf{0.238} & \underline{786.2}\\
\bottomrule
\end{tabular*}

\end{table}

\begin{figure*}[!t] \centering \includegraphics[width=\textwidth]{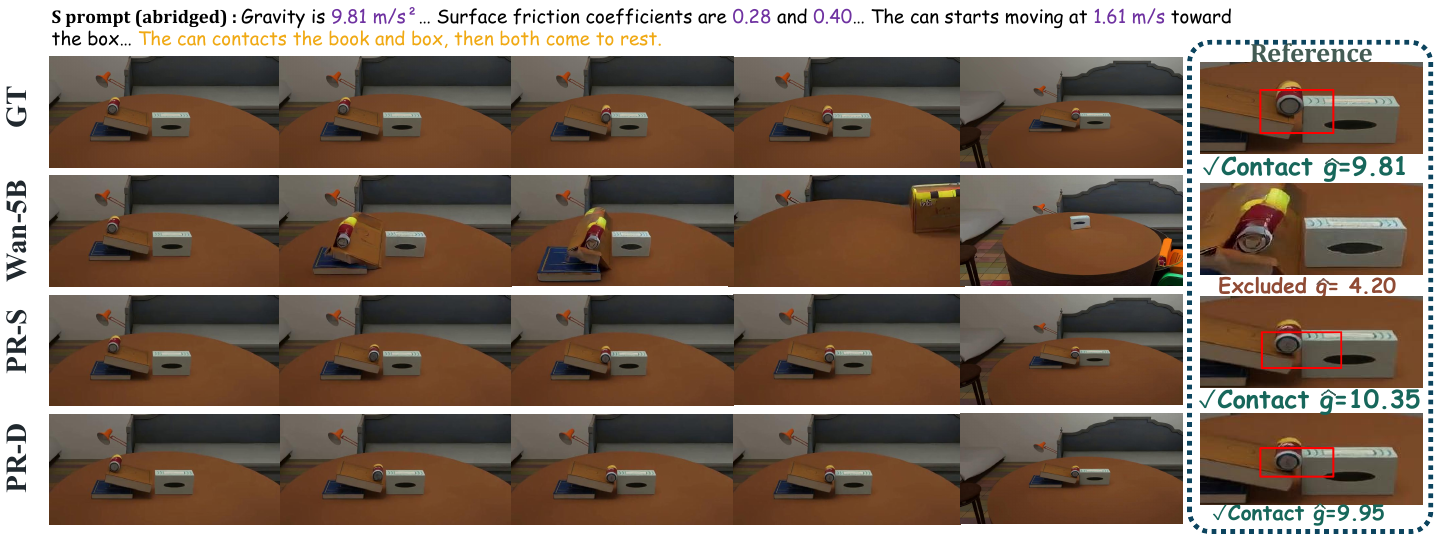} \caption{\textbf{Matched fine-tuning example.} T17\_C03, Standard prompt: GT, Wan2.2-TI2V-5B, PR-S and PR-D. Right insets show fitted gravity and contact outcomes; the Wan2.2-TI2V-5B result is marked as excluded.} \label{fig:finetune_qualitative} \end{figure*}

\subsection{Results Analysis}

\subsubsection{Parameter-level evaluation}
Even the lowest parameter errors exceed GT errors by over an order of magnitude (Table~\ref{tab:main_results}), consistently across retained shares (Supp.~Fig.~13). Recovered values also barely follow the prompts: within scenes, the rank correlation between prompted and recovered gravity, net of a grid-only null, is $-0.13$ to $+0.16$ across the ten models, against $+0.77$ for GT (Supp.~G). MiniMax-H3 leads on gravity and friction, yet under a Moon-gravity prompt its motion matches $\widehat g=11.8$\,m/s$^2$ (Fig.~\ref{fig:identifiability}). PhysRVG and VideoGPA post-train Wan2.2-TI2V-5B and PhysAlign post-trains Wan2.2-I2V-A14B, yet their gains over these bases are uneven: PhysRVG lowers friction error but raises gravity and speed errors, VideoGPA lowers gravity and friction errors but raises speed error, and PhysAlign mainly lowers speed error.

Visual similarity can reward weak motion: repeating the condition frame beats every model in PSNR, SSIM and L2 error. Among models, PhysRVG leads in PSNR and L2 error yet barely moves, and has the largest gravity error. Causal order is also poorly preserved: nine models score 1--28\% and MiniMax-H3 54\%, versus 99\% for GT.
\subsubsection{Event-level evaluation}
{\interlinepenalty=10000 Near/Far accuracy varies with model and prompt (Table~\ref{tab:rq3_strict_pairs}). The 10\%, 15\% and 20\% cutoff analysis reveals different distance responses: tightening the cutoff lowers NearAcc for Wan2.2-TI2V-5B Detailed but raises it for Cosmos3-Nano Detailed (Supp.~Table~14).\par}

Success on individual conditions does not ensure correct cross-boundary responses. Despite the highest NearAcc, 67.0\% and 69.6\%, MiniMax-H3 attains PairAcc of only 23.1\% and 18.7\%. Wan2.2-I2V-A14B instead leads in PairAcc at 30.7\% and 42.1\%, and in VPSR at 22.9\% and 34.6\%. Even the strongest model scores below 50\% PairAcc, while PhysRVG scores 0\% under both prompt formats.

Despite outcome descriptions in Standard and Detailed prompts, paired success remains limited. Under Concat prompts without descriptions of the registered outcomes, PairAcc ranges from 0 to 12.0\% across the ten models on the same 51 pairs, highlighting sensitivity to prompt formulation (Supp.~I).

\subsubsection{Complementary physical diagnoses}
\label{sec:complementarity_experiments}
Joint parameter and event evaluation reveals complementary failure modes (Supp.~Table~39). Among 2,385 gravity records evaluated at both levels, 48 have small parameter errors but incorrect event outcomes, while 929 have large parameter errors yet correct outcomes. Both patterns also occur for friction and initial speed. In Fig.~\ref{fig:bidirectional_errors}, Cosmos-Predict2 reproduces contact despite wrong speed, while Cosmos3-Nano recovers gravity but misses the domino chain.

\subsection{Fine-tuning on PhysResponse}

\label{sec:training_utility}

We fine-tune Wan2.2-TI2V-5B on 16 types, excluding benchmark cases and source-sharing videos, and evaluate 80 ID cases and 20 cases from four held-out types. PR-S/PR-D continue training Standard/Detailed text-only LoRA versions with a numerical encoder that adds sinusoidal embeddings of prompted values to their text tokens (Fig.~\ref{fig:finetune_qualitative}, Supp.~E).

Both variants lower all three parameter errors and improve motion agreement (Table~\ref{tab:training_utility}). Held-out gravity error falls from 140.3\% to 39.8--68.6\%, and speed error from 93.1\% to 13.0--16.4\% (Supp.~Table~36). Relative to their text-only versions, PR-S/PR-D further lower gravity error on the 100 cases, from 71.0\%/85.3\% to 69.1\%, and all three held-out errors (Supp.~Tables~34 and~36). PR-S improves all ten event metrics; PR-D lowers three, including Detailed PairAcc (Table~\ref{tab:ft51_comparison}). Parameter gains need not improve events.

On real WorldBench videos~\cite{worldbench}, both variants also lower gravity and ramp-friction errors (Supp.~Table~37). On 100 PhysInOne training-split scenes~\cite{physinone}, they outperform the base model and PhysInOne LoRA/SFT in PMF, PSNR, SSIM and LPIPS; SFT retains the best FVD (Table~\ref{tab:physinone_main}).

\section{Discussion and Limitations}
Effective parameters depend on context $c$, so compensating quantities such as $\mu g$ need conditional estimates (Eq.~\ref{eq:sliding_ambiguity}). Real-video transfer is tested only on WorldBench, and held-out types add no new objects or parameter ranges. Event scores rest on few scene clusters and partly on visual proxies. Encoder gains are not isolated from continued training, and benefits for embodied policies remain untested.

\section{Conclusion}
We present PhysResponse, a controlled multi-object video dataset and benchmark for evaluating physical response through parameter recovery and critical-transition events. Tests on ten models show that correct events can coexist with incorrect dynamics, while visual metrics often miss these failures. Fine-tuning reduces parameter errors on held-out interactions and transfers to external benchmarks. For video models to serve as world simulators, they must capture how changes in physical conditions alter future motion and interactions. PhysResponse provides training data and evaluation tools for this capability, a foundation for predicting action consequences in embodied planning.

\FloatBarrier
{\small
\setlength{\bibsep}{0pt plus .2pt}
\setlength{\baselineskip}{10.5pt}

}

\end{document}